\def\year{2022}\relax
\documentclass[letterpaper]{article} % DO NOT CHANGE THIS
\usepackage{aaai22}  % DO NOT CHANGE THIS
\usepackage{times}  % DO NOT CHANGE THIS
\usepackage{helvet}  % DO NOT CHANGE THIS
\usepackage{courier}  % DO NOT CHANGE THIS
\usepackage[hyphens]{url}  % DO NOT CHANGE THIS
\usepackage{graphicx} % DO NOT CHANGE THIS
\usepackage{natbib}  % DO NOT CHANGE THIS AND DO NOT ADD ANY OPTIONS TO IT
\usepackage{caption} % DO NOT CHANGE THIS AND DO NOT ADD ANY OPTIONS TO IT
\DeclareCaptionStyle{ruled}{labelfont=normalfont,labelsep=colon,strut=off} % DO NOT CHANGE THIS
\usepackage{booktabs}
\usepackage{amsmath}

\title{Deconvolutional Density Network: Modeling \\  Free-Form Conditional Distributions}
\author{
    Bing Chen$^{1,}$\equalcontrib, 
    Mazharul Islam$^{1,}$\equalcontrib, 
    Jisuo Gao$^{1}$, %\setcounter{footnote}{0}
    Lin Wang$^{1,2,}$\footnote{Corresponding author: Lin Wang}
}
\affiliations{
\textsuperscript{\rm 1} Shandong Provincial Key Laboratory of Network Based Intelligent Computing, \\University of Jinan, Jinan 250022, China\\
\textsuperscript{\rm 2} Quancheng Shandong Laboratory, Jinan 250100, China\\
\{binvecxyz, mazharul.cse34, gaojisuo, wangplanet\}@gmail.com
}

\DeclareMathOperator*{\argmin}{arg\,min}

\begin{document}

\maketitle

\begin{abstract}
Conditional density estimation (CDE) is the task of estimating the probability of an event conditioned on some inputs. A neural network (NN) can also be used to compute the output distribution for continuous-domain, which can be viewed as an extension of regression task. Nevertheless, it is difficult to explicitly approximate a distribution without knowing the information of its general form \emph{a priori}.
In order to fit an arbitrary conditional distribution, discretizing the continuous domain into bins is an effective strategy, as long as we have sufficiently narrow bins and very large data. However, collecting enough data is often hard to reach and falls far short of that ideal in many circumstances, especially in multivariate CDE for the curse of dimensionality. In this paper, we demonstrate the benefits of modeling free-form conditional distributions using a deconvolution-based neural net framework, coping with data deficiency problems in discretization. It has the advantage of being flexible but also takes advantage of the hierarchical smoothness offered by the deconvolution layers. We compare our method to a number of other density-estimation approaches and show that our Deconvolutional Density Network (DDN) outperforms the competing methods on many univariate and multivariate tasks. The code of DDN is available at \emph{https://github.com/NBICLAB/DDN}
\end{abstract}

\section{Introduction}
In supervised discriminative learning, a \emph{conditional model} learns the conditional distribution function of the outputs given the inputs, while ignoring the distribution of the inputs. If the outputs are random variables, a learned conditional model can map inputs to a discrete set of labels (e.g.\, classifying handwritten digits), or to values in a continuous domain (e.g.\ regressing temperature).

\emph{What if the expected conditional output is an explicit whole distribution, but not just a value of a random variable?} Both classification and regression can be posed as summary statistics of distribution: the location of the maximum (mode) for classification, and the expectation of the distribution for regression. However these do not tell the whole story, and often the entire  distribution is much preferred, especially when there are lots of uncertainties, and the distributions are broad or multi-modal. For instance, we can predict the distribution curve over a student's all possible scores given his personality and previous scores, or to estimate the distribution surface over all drop-off locations given pickup location and the time of day \cite{GP_CDE}. 
\begin{figure}
    \centering
    \includegraphics{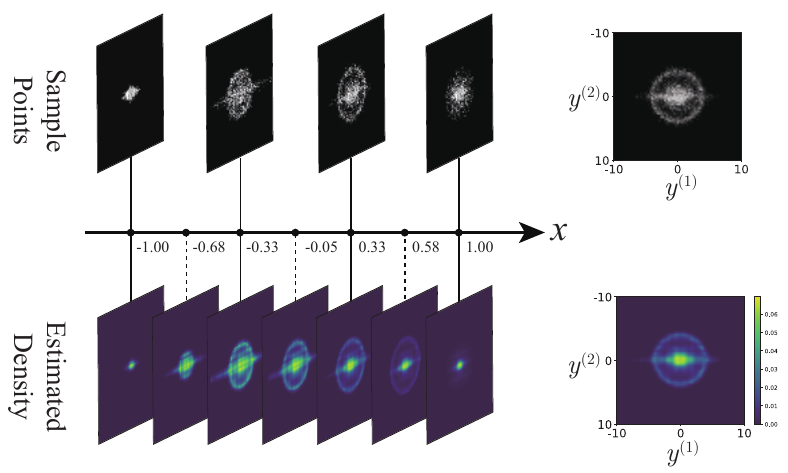}
    \caption{Example Conditional Density Estimation, estimating the conditional density on a 2-dimensional domain. This dataset has 35,000 training samples in total, drawn from a distribution constructed based on an animation, by taking each  normalized frame and time as conditional density function and condition, respectively. Each sample point consists of a single input $x$ and two outputs $(y^{(1)}, y^{(2)})$. We wish to estimate the conditional density over all possible  values of $(y^{(1)}, y^{(2)})$, \textbf{given different input/condition $x$}. Given different value of input, the sub-figures at top and at bottom represent the scatter plot of training samples and the estimated conditional densities, respectively. Notice that the density can change markedly, depending on the input. The density is estimated by DDN.}
    \label{fig:schematic}
\end{figure}

This is related to the Conditional Density Estimation (CDE), a more general discriminative learning task, where the goal is to return the distribution over a range of output values. More precisely, given an input $x$, estimate $p (y|x)$, where $y \in \mathcal{D}$ and $\mathcal{D}$ stands for the domain of $y$.  Fig.~\ref{fig:schematic} illustrates an example of CDE.

We focus on \emph{explicit} CDE for fast and direct access to the likelihoods and the whole landscape of distribution. A critical problem with learning an explicit CDE model is the ground truth distribution as training signal is often absent. Instead, we can maximize the likelihood. For example, if $y$ belongs to discrete-domain, one could take a neural network classifier using softmax as output layer and cross-entropy as loss function to learn an explicit estimation.

However, if $y$ belongs to continuous-domain, constructing a distribution for explicit CDE tasks becomes not that straightforward. In theory, an infinite number of control parameters is needed to specify arbitrary probability density functions (PDFs) at continuous-domain. This is not practically possible, and is further exacerbated by the finite size of datasets used to tune the parameters.

The only option is to limit ourselves to a finite-dimensional subspace for our output functions. For example, the model could learn to output the mean and standard deviation of a Gaussian distribution. Or, we could output a number of such combinations of parameters, constructing a PDF as a Gaussian mixture model \cite{BISHOP_2006} or as normalizing flows \cite{Trippe_2018}.  

Hence, we need to apply some assumptions, so-called \emph{inductive biases}, to the CDE model for continuous-domain. Naturally, it is helpful if one knows the general form of the target distribution \emph{a priori}. However, there are many cases in which that information is not available, and our chosen form of inductive bias might not fit the underlying distribution very well. Therefore, considering the distribution diversity, we would like to make as few assumptions about the distribution as possible to construct a free-form PDF.

One effective way to construct such free-form PDF curve is \emph{piecewise-constant function}, in which we specify the values of the function over a set of discrete bins. Ideally, \emph{if each bin is narrow enough and has sufficient training samples to draw from}, then the law of large numbers should yield a piecewise-constant function that faithfully reflects the underlying true distribution. Despite being a parametric function, this provides extensive flexibility. Moreover, compared to mixture density models, it is easy to learn and produces better performance, as it avoids parts of PDF moving beyond a predefined region \cite{PixelRNN}. In fact, the effectiveness of piecewise-constant function for constructing free-form distribution has been verified in generative learning, like PixelRNN \cite{PixelRNN} and  Gated PixelCNN \cite{GatedPixelCNN}.

\subsubsection{Motivation}
Unfortunately, the realities of dataset sizes fall far short of that ideal in many circumstances, especially when we only have \emph{few data samples}. Actually, collecting enough data is hard to reach for many fields in many circumstances. In addition, if we want to finely approximate a conditional density function using many narrow bins \cite{LI_2021}, the data deficiency problem would also lead to poor performance. This problem is even more notable for multivariate CDE on account of the curse of dimensionality, especially when $x$ is in continuous-domain.

Therefore, a question arises here is: \emph{can we construct a piecewise-constant function to approximate free-form PDF, on the condition that only few samples are available for learning?}

\subsubsection{Contribution} The answer is yes. The shortage of data in each bin can be mitigated by applying a hierarchical smoothness constraint, which is the main objective of this research. 

In this paper, a Deconvolutional Density Network (DDN) is proposed to estimate free-form explicit conditional distributions. It adopts deconvolution \cite{DCGAN_YU} (including forms of convolution with upsampling \cite{Shelhamer_Cov_upsam}) to naturally encourage smoothness because of its built-in hierarchical spatial correlation properties. 
The contribution is decomposed as follows:
\begin{enumerate}
\item This work designs a deconvolution-based neural net framework for CDE to explicitly estimate free-form PDF for continuous-domain, especially coping with the data deficiency problem.
\item Experiments manifest DDN achieves the best results on diversified univariate and multivariate CDE tasks in comparison with six different explicit CDE models.
\end{enumerate}

\section{Related Work}

This section reviews previous work on CDE and describes the motivations behind our method. 

Conceivably kernel-based methods are among the most popular non-neural CDE techniques, as they involve combining kernel functions with data to calculate the probability distributions by interpolating between observed points to predict the probability density of the unseen points. One of the well-recognized kernel-based approaches to estimate CDE is to use the kernel density estimator (KDE) \cite{Tsybakov_2008}. In principle, KDE can approximate any probability distribution without any prior assumptions. The kernel quantile regression (KQR) method can predict the percentiles of the conditional distribution \cite{Takeuchi_2006, Li_2007, Takeuchi_2009}. A straightforward CDE method, least-squares conditional density estimation (LSCDE), was proposed by \citeauthor{Sugiyama_2010}, employing the squared loss function.

Neural networks have been used for CDE as a result of their expressive power. Mixture density networks (MDNs) \cite{BISHOP_2006} are among the earliest strategies for modeling conditional densities as a mixture of distributions, employing NNs to estimate the parameters of each distribution (e.g. mean, standard deviation, etc.). To solve the quadratically growing problem of parameter number in MDNs, the real-valued neural autoregressive density-estimator (RNADE) applies one-dimensional MDNs to unsupervised density estimation in a practical way \cite{RNADE}.

Another big family of neural density estimators that are flexible is normalizing flows \cite{cms/1266935020, Tabak2013AFO, Rezende_2015}, involving applying different invertible transformations to a simple density until a desired level of complexity is reached. They provide the PDF with a tractable form. In terms of estimating conditional density, \citeauthor{Trippe_2018} employ normalising flows as a flexible likelihood model to fit complex conditional densities \cite{Trippe_2018}. 

A model agnostic normalizing flow based noise regularization method to estimate conditional density was proposed by \citeauthor{rothfuss2020noise}, which controls random perturbations to the data during training \cite{rothfuss2020noise}. Utilizing normalizing flow to model conditional distributions, the inverse problems employing maximization of the posterior likelihood is investigated \cite{xiao2019method}. The Conditional Normalizing Flows (CNFs), where the base models to output space mapping is conditioned on an input, is presented by \citeauthor{CNFs} to model conditional densities. A normalizing flow based autoregressive model \cite{MAF} was proposed for density estimation, by stacking multiple models with similar characteristics. 

The study of latent space models has also generated considerable interest in CDE over the past few years. The Gaussian Process CDE \cite{GP_CDE} uses latent variables to extend input and maps augmented input onto samples from the conditional distribution by Gaussian processes. As a variant of variational autoencoder \cite{Kingma_2019_VAE}, the conditional variational autoencoder (CVAE) is another way to approach CDE \cite{CVAE}.

There are also other miscellaneous neural network techniques for CDE. The kernel mixture network (KMN) \cite{Ambrogioni_2017} combines non-parametric and parametric elements by formulating the CDE model as a linear combination of a family of kernels. On the basis of score matching, \citeauthor{sasaki_2018} designed a neural-kernelized conditional density estimator (NKC) to estimate the conditional density. A conditional sum-product networks (CSPNs) is proposed \cite{Conditional_Sumproduct_Networks} for multivariate and potentially hybrid domains that allow harnessing the expressive power of neural networks while still maintaining tractability guarantees. A neural noise-regularization and data-normalization scheme for CDE was used for financial applications, addressing problems like overfitting, weight initialization, and hyperparameter sensitivity \cite{Jonas_2019}.  \citeauthor{strauss2021arbitrary} devised an energy-based approach for conditional density estimation to eliminate restrictive bias 
from standard alternatives \cite{strauss2021arbitrary}.

If we move our focus to \emph{generative learning} temporally, there are some flexible approaches transforming the continuous-domain density estimation tasks to a discrete domain by piecewise-constant function. The PixelRNNs \cite{PixelRNN} sequentially predicts the pixels in an image by treating the pixel values as discrete random variables using a softmax layer in the conditional distributions. A PixelCNN is also proposed in that work, sharing the same core components as the PixelRNN. Based on PixelRNN, a computationally efficient variant Gated PixelCNN and its conditional version are also proposed \cite{GatedPixelCNN} to model the complex conditional distributions of natural images. The PixelCNN++ \cite{PixelCNN++} improves PixelCNN by a discretized logistic mixture likelihood on the pixels together with other strategies to speed up training and simplify the model structure.

Let's get back to \emph{deterministic learning} now. Recently, there have been some preliminary attempts on discretizing CDE for continuous-domain discriminative learning. A multiscale network transforming the CDE regression task into a hierarchical classification task was proposed in \cite{Tansey_2016} by decomposing the density into a series of half-spaces and learning Boolean probabilities for each split. Ref.\cite{LI_2021} described a conditional distribution technique, named as joint binary cross entropy (JBCENN), based on deep distribution regression, redefining the problem as many binary classification tasks. The neural spline flow model (NSF) \cite{Neural_Spline_Flows} leverages piecewise densities defined over a set of bins.

\subsubsection{Challenge}
While kernel-based methods are effective with a limited number of random variables, they scale poorly as the number of variables increases. The MDN and normalizing flow can theoretically approximate any distribution, but if the form of the target distribution differs significantly from the basis functions (e.g. Gaussian or Radial Flows), it is still difficult to accurately learn a continuous distribution. Moreover, the latent space model, like CVAE, usually adopts Monte Carlo (MC) sampling to estimate likelihood, which is very time-consuming and generally not explicit.

\begin{figure*}
\centering
\includegraphics{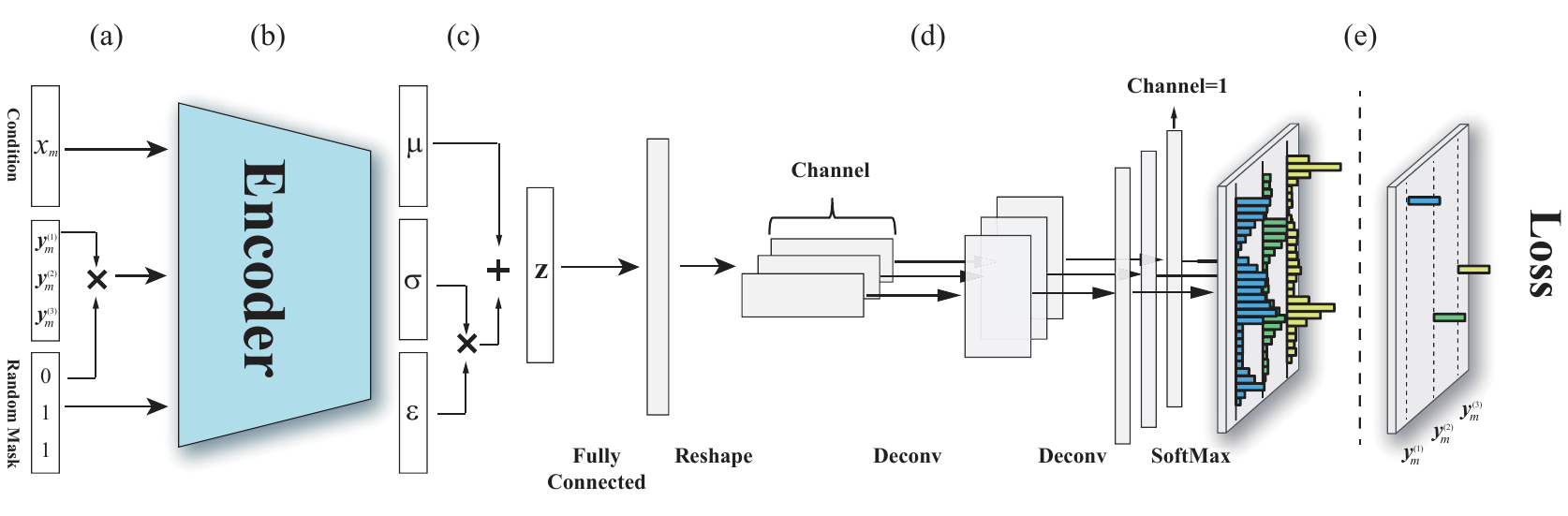}
\caption{The framework of deconvolutional density network (DDN). The framework consists of five successive parts: (a) a preprocessing module for regulating input conditions, (b) an encoder for mapping the inputs to a latent space, (c) a variational layer with optional stochasticity, (d) a multi-layer deconvolutional estimator for hierarchical topological smoothness, followed by (e) a partitioning-based loss. Note that it only takes one dimensional deconvolutional layers at (d) for each component $y_m^{(j)}$ of a sample $(x_m, y_m)$ independently, but not implies two-dimensional kernels. In addition, ${\epsilon} {\sim} \mathcal{N} (0,I)$ during training, whereas the $\epsilon$ is set to zeros during testing to guarantee that a density is tractable. According to the random mask $(0, 1, 1)$ for $y_m$ at (a), the DDN is predicting three univariate distributions $p(y_m^{(1)}|x_m,y_m^{(2)},y_m^{(3)})$, $p(y_m^{(2)}|x_m,y_m^{(2)},y_m^{(3)})$, and $p(y_m^{(3)}|x_m,y_m^{(2)},y_m^{(3)})$. }
\label{fig:model}
\end{figure*}

According to PixelRNN \cite{PixelRNN}, JBCENN \cite{LI_2021}, and NSF \cite{Neural_Spline_Flows}, using a piecewise-constant function to discretize the continuous-domain distribution has the potential to approximate a free-form distribution, considering we have sufficiently narrow bins, and very large data. Actually, \citeauthor{PixelRNN} experimentally found the piecewise-constant function-based CDE is ``easy to learn and to produce better performance compared to a continuous distribution"\cite{PixelRNN}. 

But the strain between infinite-dimensional distributions and finite datasets makes it difficult to faithfully estimate a distribution by piecewise-constant function, especially in scenarios when we only have a small number of samples. 
This would produce distributions that can be very non-smooth. Furthermore, the absence of hierarchical smoothness in these methods may also result in failure when predicting complex distribution on a small training set. In addition, the PixelRNN \cite{PixelRNN} and JBCENN \cite{LI_2021} do not investigate on the exponential computational cost problem when using a piecewise-constant function for multivariate tasks.

In DDN, assuming the true PDFs are commonly smooth and have few spikes, we can use multi-layer deconvolution to hierarchically construct distributions that exhibit spatial coherence at multiple scales, coping with data deficiency problems. Furthermore, a conditional masking-based univariate distribution estimation strategy is adopted to construct an auto-regressive multivariate CDE model. In the next section, we describe how our Deconvolutional Density Network (DDN) can infer free-form density functions.

\section{Methodology}
\label{sec_method}

The complete framework of the proposed Deconvolutional Density Network (DDN) is shown in Fig.~\ref{fig:model}, combining five successive modules into a single framework. This framework enables DDN to flexibly and explicitly estimate conditional density functions for continuous-domain, reducing needs to large size data. We will describe how the DDN works from Fig. \ref{fig:model}(e) back to Fig. \ref{fig:model}(a), subsection by subsection.

\subsection{Baseline Model}
\label{subsec_baseline_model}
\emph{Here we briefly describe the baseline model, and other components of DDN will be added into it progressively.}

If we are given a set of $M$ input/output samples $(x_m, y_m)$, then the likelihood of observing those samples under our model is,
$$
\prod_{m=1}^M g \left( y_m ; f(x_m, \theta) \right) .
$$
Taking the log of that likelihood, and negating it, yields the negative log-likelihood,
$$
- \sum_{m=1}^{M} \log g \left( y_m ; f(x_m, \theta) \right) \ .
$$
Our NN model can be trained by attempting to minimize the expected negative log-likelihood of the training data,

\begin{equation}
\small
\label{eqn:minimize_nll}
\begin{split}
\argmin_\theta E_{(x_m, y_m)} \Big[ - \log g(y_m; f(x_m, \theta)) \Big] \quad  
& =\\ \quad \argmin_\theta \left[ - \frac1M \sum_{m=1}^M \log g(y_m; f(x_m, \theta) ) \right]\ .\end{split}
\end{equation}

\subsection{Partitioning}
\label{subsec_partitioning}
Partitioning is the first step  for discretizing PDF of $y$ of continuous-domain (see Fig. \ref{fig:model}(e)). In this subsection, we start from the univarite task, i.e., $y$ is one dimensional. We assume $\mathcal{D}$, the domain of the variable $y$, is finite and partitioned into a number of uniform bins. These bins will be used to construct the CDE as a piecewise-constant function. Discretizing $g$, we partition the output space into $N$ bins, and denote the $i^\mathrm{th}$ bin using ${B}_{i}$, with uniform bin width $\Delta B$. If our neural network outputs $N$ values, $f_i$ for $i=1, \ldots , N$, then we simply interpret $f_i(x, \theta)$ as the estimated probability that $y$ falls in the $i^\mathrm{th}$ bin. Then, the estimated conditional density function can be expressed as a piecewise-constant function,
$$
g(y \mid x) = \left\{
\begin{array}{cl}
    \vdots & \\
    \frac{f_i(x, \theta)}{\Delta B} & \mathrm{if} \ y \in B_i \\
    \vdots
\end{array}
\right. .
$$
We can write the negative log-likelihood loss function as,
\begin{align}
\small
& \argmin_\theta \left[ - \frac1M \sum_{m=1}^M \log g(y_m \mid x_m ) \right] \\
=& \argmin_\theta \left[ - \frac1M \sum_{m=1}^M \sum_{i=1}^{N}\log \left( f_i \left(x_{m}, \theta \right) \right) \mathbf{1}_{B_i}\left(y_{m}\right)\right], \label{eqn:3}
\end{align}
where ${\bf 1}_{B_i} (y)$ is an indicator function, equal to 1 when $y \in B_i$, and zero otherwise.

Let's move to the multivariate CDE tasks now, i.e., $y$ is $J$-dimensional. In DDN, we translate a $J$-dimensional CDE task into $J$ one-dimensional CDE tasks at first, and take the chain-rule of probability \cite{Trippe_2018, strauss2021arbitrary} to compose them as
\begin{equation}
\label{eqn:chain_rule}
\begin{array}{l}
p\left( {{y^{(1)}},{\rm{ }} \cdots ,{y^{(J)}}|x} \right) = \prod\limits_{j = 1}^J {p\left( {{y^{(j)}}|x,{y^{(1)}},{\rm{ }} \cdots ,{y^{(j - 1)}}} \right)}
\end{array}.
\end{equation}
Suppose the network has $J$ different sets of output nodes, where the number of nodes in each set is $N$, corresponding to $y^{(j)}$, the $j^\mathrm{th}$ variable of $y$. Fig.~\ref{fig:model} shows three such sets of outputs. Then we can include all of these outputs in our loss function from Eq. \ref{eqn:3}, avaraging over them all. Thus, our loss function for multivariate tasks includes a new average over $J$,
\begin{equation}
\small
\label{eqn:final_nll_loss}
\argmin_\theta \left[ { - \frac{1}{M}\frac{1}{J}\sum\limits_{m = 1}^M {\sum\limits_{j = 1}^J {\sum\limits_{i = 1}^N {\log } } } \left( {{f_{ij}}\left( {{x_m},\theta } \right)} \right){{\bf{1}}_{{B_{ij}}}}\left( {y_m^{\left( j \right)}} \right)} \right],
\end{equation}
where $f_{ij}(x, \theta)$ is the output of the $i^\mathrm{th}$ node in the $j^\mathrm{th}$ set of output nodes, and ${\bf 1}_{B_{ij}}(y^{(j)})$ is the indicator variable that equals 1 when $y^{(j)} \in B_{ij}$, and zero otherwise.

\subsection{Deconvolutional Estimator}
\label{subsec_deconv_estimator}

In this subsection, we focus on Fig. \ref{fig:model}(d), which is the critical part of our deconvolutional density network. Despite its potential to approximate a free-form conditional distribution, discretizing the continuous-domain distribution by a piecewise-constant function often suffers from statistical undersampling in its bins, resulting in a distribution that is very spiky. Moreover, the conditional density in real-world can be multi-modal and very complex.

\emph{If we can assume that the true PDFs are commonly smooth and have few spikes, we could include a hierarchical mechanism that correlates nearby outputs. } In this case, hierarchical deconvolution is a strong candidate for modeling complex landscape, as it can introduce smoothness and structural details layer by layer.

In this work, a multi-layer deconvolutional estimator $h$ is designed to instill spatial coherence at different scales. This has the advantage of constructing the PDF in a multiscaled, hierarchical manner through multiple deconvolution layers, allowing the model to introduce structure and smoothness at different scales.

Note that our deconvolutional layers are not there to invert or undo a convolutional layer. Instead, the deconvolutional layers \cite{DCGAN_YU} (also known as transposed convolution or upsampling convolution \cite{Shelhamer_Cov_upsam}) are used to build the output layer-by-layer, in progressively higher resolution.

In our deconvolutional estimator $h$, the latent vector $z$ (see next subsection) is firstly mapped to a fully connected layer, which is further reshaped into a multi-channel feature map. These feature maps are split into $J$ groups, each of which corresponds to a variable of $y$. Note that, each variable of $y$ has an independent pathway from the reshaped initial feature map to its own distribution.

The reshaped initial feature maps are then fed into a sequence of deconvolution layers, constructed by upsampling and convolution. Upsampling expands the feature map by an integer scale factor, while the convolution transforms the expanded feature map with shared weights introducing spatial correlation. After these deconvolution layers, we obtain a vector of unnormalized logits, which are finally fed into a SoftMax layer to get a discretized probability vectors. At this point, we have $J$ discretized probability vectors, which are used in the loss functions (\ref{eqn:final_nll_loss}). 

\subsection{Encoder-Estimator Framework}
\label{subsec_encoder_estimator_framework}

Let's move our attention to Fig. \ref{fig:model} (b \& c) now. Although data insufficiency problem in piecewise-constant function can be mitigated by deconvolution. There are still some remaining questions.

\emph{What if we only have extremely small number of samples for learning? What if the target PDF is not that smooth? How to calibrate the estimated distribution?}

We wish the DDN to focus on the statistically salient features of limited data, not the contingencies specific to our particular training samples. 

Since the  layer by layer abstraction enables neural network to extract underlying features of data, adding noise to each sample at latent space could provide the model with lots of unseen but reasonable patterns during learning, to solve the data deficiency problem.

Therefore, we split our network into two main parts: an encoder $q(z | x)$ (see Fig. \ref{fig:model}(b)), and an estimator $h(y|z)$; a latent space (see Fig. \ref{fig:model}(c)) sits between the two.

This is inspired by the Deep VIB framework \cite{Alemi2017, Alemi_calibration}. The encoder outputs two equal-sized vectors, $\mu$ and $\sigma$, representing the mean and standard deviation for a multivariate, diagonal Gaussian distribution. The latent vectors $z$ are stochastic, drawn from that Gaussian distribution using the reparameterization trick, $z =\mu + \sigma \cdot {\epsilon}$, where  ${\epsilon}$ is drawn from the standard Normal distribution, ${\epsilon} {\sim} \mathcal{N} (0,I)$. In order to seek a less complex and more economical latent representation, we could force the network to sacrifice some encoding fidelity \cite{Alemi2017} by adding KL-divergence term $L_{\mathrm{KL}}$ to loss function, where $L_{\mathrm{KL}}= D_\mathrm{KL} \big[ q\left(z| x\right) \big\| \, \mathcal{N} (0,I) \big]$ and $q(z|x) \sim \mathcal{N} (\mu(x; \theta),\sigma ^2(x; \theta))$. In DDN,  we call this latent-space layer a \textit{variational layer} (VL).

The adoption of Encoder-Estimator framework makes the final objective function of DDN be written as
\begin{equation}
\label{eqn:final_loss}
\small
\begin{split}
\frac{1}M  \sum_{m=1}^M \left\{E_{\epsilon \sim N(0,I)}\left[ -\frac{1}{J} \sum_{j=1}^{J}\sum_{i=1}^{N}\log \left( h_{ij}\left(y_{m}^{(j)} | z \right)\right) \right.\right.\\\left.\left. \times \mathbf{1}_{B_{ij}}\left(y_{m}^{(j)}\right) \Bigg. \right] + \beta L_{\mathrm{KL}} \Bigg. \right\}.
\end{split}
\end{equation}
Recall that $h_{ij}$ and $q$ are both functions of the network parameters, $\theta$. The $\beta$ coefficient influences on the \emph{calibration of estimated distribution} of DDN directly \cite{Alemi2017,Higgins2017betaVAELB}. When $\beta=0$, the latent-space distribution can relax, and spread out, allowing the samples to disambiguate from each other. Increasing $\beta$ encourages the latent-space distribution to contract toward the origin \cite{Alemi2017}. In this way, $\beta$ influences the calibrating estimated distribution \cite{Alemi_calibration} by forcing the model to focus on the statistically salient features, and thereby help to improve the confidence.

Notice that, different from CVAE \cite{CVAE} or other related latent space CDE models, the DDN doesn't need Monte Carlo sampling on $z$ to access $p(y|x)$. In addition, one may argue that we may have to integrate over $z$ to obtain an explicit expression for $p(y|x)$, which is generally intractable. Instead, during testing of DDN, we simply pass the mean vector to the deconvolutional estimator $q$. In practice, this is realized by setting $\epsilon$ to 0 to remove latent-space stochasticity.

\subsection{Conditional Masking}
This subsection is about Fig. \ref{fig:model}(a). The DDN relies on partitioning the domain of $y$ into uniform bins, which has a computational cost exponential in $J$, where $J$ is the dimension of $y$. For multivariate tasks, in order to avoid the number of bins produce scales exponentially with the dimension, we adopt auto-regressive structure \cite{pmlr-v15-larochelle11a} using the chain-rule of probability (recall subsection Partitioning). Given this, \emph{how to calculate the univariate distributions in chain-rule of probability?}

In this work, we adopt a conditional masking mechanism to configure DDN for predicting different univariate conditional densities, which can be further assembled for the target multivariate density. At the beginning of learning, we first generate $K$ permutations for $(1, 2, \hdots, J)$. Given the $k^\mathrm{th}$ permutation is defined as $(A_1^k, A_2^k, \hdots, A_J^k)$, according to Eq. (\ref{eqn:chain_rule}), the $k^\mathrm{th}$ path to calculate the multivariate conditional distribution is

\begin{equation}
\label{eqn:chain_rule_perm}
\begin{split}
&p\left( {{y^{(1)}},{y^{(2)}}, \cdots ,{y^{(J)}}|x} \right) \\
&= \prod\limits_{j = 1}^J {p\left( {{y^{(A_j^k)}}|x,{y^{(A_1^k)}}, \cdots ,{y^{(A_{j - 1}^k)}}} \right)}.
\end{split}
\end{equation}

We further define a conditional mask vector set $S$ for masking $y$, each element of which is a $J$-dimensional binary vector. Each mask vector corresponds to a condition in a path, i.e., $(y^{(A_1^k)}, \hdots ,y^{(A_{j-1}^k)})$ except $x$. Note the duplicate conditions in $S$ should be reduced. For a condition and its corresponding mask vector, the $j^\mathrm{th}$ component of mask is set to one if $y^{(j)}$ belongs to this condition, otherwise it is set to zero.

During learning, given a sample $(x_m,y_m)$, we randomly choose a mask $rm$ from $S$ at first, and concatenate $x_m$, $rm$, and $y_m \cdot rm$ as the input to DDN. In this way, we can easily assemble different univariate conditional models into a single DDN framework. For example, a conditional binary mask $(1, 0, 1)$ for $y_m$ means we want to set $(x_m, y_m^{(1)},y_m^{(3)})$ as a condition. 

The mask actually plays a role of information gate, i.e., one means letting it in and zero means keeping it out. Particularly, when $y_m \cdot rm$ is zero, mask $rm$ is one means the zero comes from domain. Otherwise, if mask $rm$ is zero, the information of $y_m$ is blocked.

In test, we continue to take this conditional masking mechanism on DDN to access different univariate distributions and send them into Eq. (\ref{eqn:chain_rule_perm}) to calculate the multivariate conditional distribution. In addition, we freeze chain-rule paths, i.e. the mask vector set $S$, at the inference to ensure a normalized distribution. Ideally, the predicted distribution by different paths should be the same as each other. However, in practice, we take the average joint distribution from all $K$ paths to reduce predicting errors.

\section{Experiments}

This section examines the efficacy of DDN for explicitly estimating free-form conditional PDF on continuous-domain. The following questions are intended to be addressed through our experiments: 
(\textit{a}) What is the influence of each of the main components of DDN on the model's performance?
(\textit{b}) What is the influence of the key hyperparameter $\beta$ on the estimated density?
(\textit{c}) In light of the ground truth distributions, how does DDN compare to competitive techniques on low-dimensional toy tasks?
(\textit{d}) In real-world tasks, what is DDN's performance compared to the competitors?

In response to these questions, we first designed ablation experiments to demonstrate the need for different components of the proposed DDN, followed by an analysis of the hyperparameter $\beta$. Then we performed experiments on four toy tasks with known ground truth conditional distributions \(p(y|x)\) and evaluated the model using several univariate and multivariate benchmark datasets from real-world applications.

\subsection{Experimental Configuration}
\label{subsec:configuration}

\subsubsection{Toy \& Real World Tasks}
Considering the ground truth conditional distributions are important for evaluating the accuracy, we designed four 2D toy tasks which \(p(y|x)\) can be easily calculated (See Fig. \ref{fig:2D_SSE_final}), where $x \sim U(-1, 1)$. The details of generating training samples are as follows:

\begin{itemize}
    \item Squares: For given $x$; $\lambda \sim \mathrm{Bern}(0.5)$; $a^{(1)}, a^{(2)} \stackrel{iid}{\sim} U(-5 + x, -1 + x)$; $b^{(1)}, b^{(2)} \stackrel{iid}{\sim} U(1 - x, 5 - x)$;
    
    $y^{(1)} = \lambda * a^{(1)} + (1 - \lambda) * b^{(1)}$; $y^{(2)} = \lambda * a^{(2)} + (1 - \lambda) * b^{(2)}$ 
    \\
    \item Half Gaussian: For given $x$; $a, b \stackrel{iid}{\sim} N(0, 2)$; $y^{(1)} = \left|a\right| \cos x\pi - b\sin x\pi$;
    $y^{(2)} = \left|a\right| \sin x\pi + b\cos x\pi$
    \\
    \item Gaussian Stick: For given $x$; $a \sim N(0, 1)$;$b \sim U(-6, 6)$; $c = (-0.75 + x) / 2$; $y^{(1)} = a \cos c\pi - b \sin c \pi$; $y^{(2)} = a \sin c\pi + b \cos c\pi$
    \\
    \item Elastic Ring: For given $x$; $d \sim U(0, 2)$; $\theta \sim U(0, 2\pi)$; $y^{(1)} = (4 + 2 * x + d) \cos \theta$; $y^{(2)} = (4 - 2 * x + d) * \sin \theta$
\end{itemize}

The validation also included seven real-world datasets from the UCI machine learning repository \cite{Dua:2019}. We standardized all inputs and targets using z-score normalization. Table \ref{table:dataset} shows the characteristics of each dataset. All experiments were running on a Tesla V100 GPU. We used Adam optimizer with the default configuration. The learning rate was set to 3e-4, and the batch size was 256. 

\begin{table}
  \centering
  \begin{tabular}{lcccccc}
    \toprule
          & Samples & Features & Targets     \\
    \midrule
    Fish          & 908     & 6       & 1        \\
    Concrete      & 1030    & 8       & 1      \\
    Energy        & 768     & 8       & 2       \\
    Parkinsons    & 5875    & 16      & 2       \\
    Temperature   & 7588    & 21      & 2       \\
    Air           & 8891    & 10      & 3       \\
    Skillcraft    & 3338    & 15      & 4       \\
    \bottomrule
  \end{tabular}
  \caption{Characteristics of UCI datasets.}
  \label{table:dataset}
\end{table}

\subsubsection{Models} 

In all experiments, the estimator of our DDN contained two Upsample-Conv-BatchNorm-LeakyReLU blocks and one Upsample-Conv block in each target dimension. We configured the blocks so that the output space was partitioned into $N$ = 256 bins. We set the number of permutations $K$ to the maximum number of permutations $K_{max}$ if $K_{max}$ does not exceed 5; otherwise, $K$ was set to 5. We used 16 latent codes in VL. For 1D tasks, DDN is not using the masking strategy, which means that its input only condition $x$.

We conducted experiments to compare DDN with seven other neural-network-based CDE methods, including DDN without VL, MDN \cite{BISHOP_2006} employing 20 Gaussian kernels, MAF \cite{MAF} using 5 autoregressive transformations, NSF \cite{Neural_Spline_Flows} adopting autoregressive model using 5 RQS transformations, RNF utilizing 5 radial transformations \cite{Trippe_2018}, JBCENN \cite{LI_2021} using 256 bins similar to our DDN model, and Multilayer Perceptron with Softmax (MLP). 
For each model, most hyperparameters were determined by recommendation, but the ones which didn't work well on given tasks were tuned by trial and error.

All models began with the same basic three-layers network (input layer and two hidden layers) with 64 tanh neurons in each hidden layer. We normalized every hidden layer using batch-normalization. To balance the proportion of conditional information and masked information on two-dimensional and higher-dimensional tasks for our DDN, we used two independent linear layers with 32 tanh neurons to replace the first layer in the basic network, which map $x$ and $[y \cdot m, m]$ to vectors with same length 32, where $x$ is the condition, $y$ is the target value, $m$ is the mask used in DDN, and $[\cdot, \cdot]$ denotes the concatenation operation. 

It is necessary to set the range of target variables when using DDN, JBCENN, and MLP. We set the range to (-10, 10) for the toy tasks, . For real-world tasks, the range was set to $(\lfloor y_\mathrm{min} \rfloor - 1, \lceil y_\mathrm{max} \rceil + 1)$, where $y_\mathrm{min}$ and $y_\mathrm{max}$ denote the minimum and maximum values of the target random variable $y$, respectively. We set the probability density to 0 for $y$-values outside of that range for these three methods. 

MAF in this paper was constructed as described in \cite{MAF}. The basic network with an additional linear layer was used to parameterize the transformation function $z^{(i)} = f(\mathbf{x}, y^{(i)}) = (y^{(i)} - \mu(\mathbf{x}, \mathbf{y}^{(1:i-1)})) / exp(\alpha(\mathbf{x}, \mathbf{y}^{(1:i-1)})),$ where $\mathbf{x}$ is the condition and $\mathbf{y}$ is the target. Note that for one N-dimensional CDE task, a MAF was parameterized by $N$ basic networks. 

NSF was constructed using the same framework as the MAF, whose transformation function was $z^{(i)} = f(\mathbf{x}, y^{(i)}) = RQS_{\theta(\mathbf{x}, \mathbf{y}^{(1:i-1)})}(y^{(i)})$, where RQS is the rational-quadratic spline function from \cite{Neural_Spline_Flows}. For all toy tasks, we set the interval to $(-10, 10)$ for the RQS function; for all UCI tasks, we set it to $(-3, 3)$. 

The RNF framework was the same as MAF, but its transformation function was replaced by the radial function in \cite{Trippe_2018}.

\begin{figure}
\centering
\includegraphics{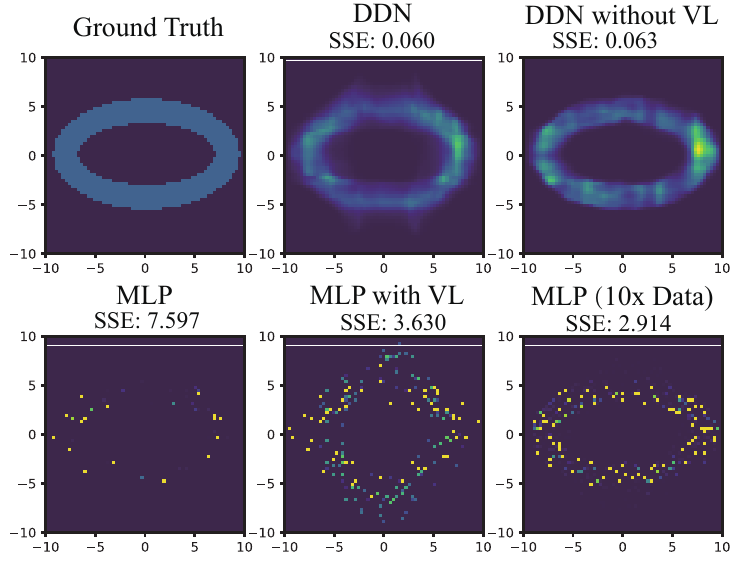}

\caption{Comparison of DDN and MLP by ablating different components. Although the MLP captures more information about the true density by coupling with VL or training on 10x samples, the estimated density is still very spiky on account of the statistical undersampling in each bin. The DDN mitigates this problem by introducing spatial correlation.
}
\label{fig:Ablation}
\end{figure}

\subsubsection{Evaluation Metrics} For real-world tasks, we used the log-likelihood (LL) to evaluate a CDE model. Moreover, we reported the standard deviation of the log-likelihood to demonstrate the stability of DDN in comparison to other techniques. Nevertheless, for the toy tasks, it's feasible to directly measure the difference between the estimated distribution and the ground truth, as the ground truth distribution \(p(y|x)\) can be easily derived.  We used the sum of squared errors (SSE) for measuring the difference.

\begin{figure}
    \centering
    \includegraphics{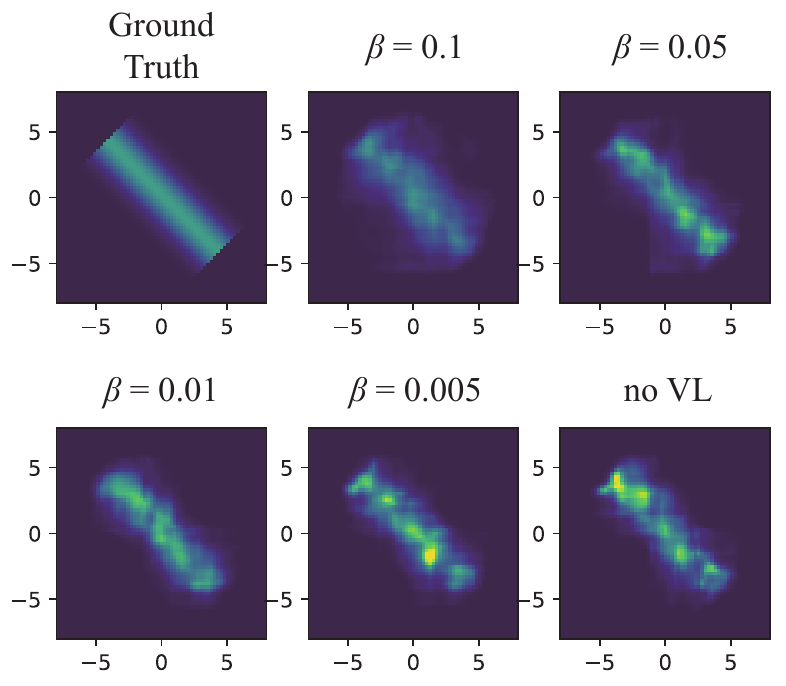} 
    \caption{Comparison of estimated densities of DDNs with different $\beta$ values on Gaussian Stick task, where the condition $x = -0.75$. }
    \label{fig:beta}
\end{figure}

\subsection{Ablation Experiment}
\label{subsec_ablation_exp}
We designed the ablation experiments on Elastic Ring task by comparing DDN, DDN without the variational layer, MLP, MLP with variational layer, and MLP with 10x data samples in estimating free-form conditional distributions. The DDN without VL was designed to show the influence of VL, as we replaced VL with an activated dense layer. Moreover, the MLP, estimating a distribution by piecewise-constant function but without deconvolution, was used to show the influence of the deconvolutional estimator. The estimated densities when $x = 0.75$ are shown in Fig.~\ref{fig:Ablation}.

MLP can capture adequate information about the true conditional distribution, only if statistically enough samples are provided in each bin. Although coupling with VL or training on 10x samples can mitigate this problem, the estimated distribution is still spiky. The problem can be exacerbated with the increase of dimension due to the curse of dimensionality. VL exposes MLP at more unseen patterns in latent space during learning, but it cannot guarantee the smoothness of density. By contrast, the DDN mitigates this problem by employing deconvolution to introduce spatial correlation hierarchically. Furthermore, note that VL alleviates the over-fitting problem of DDN, and enables calibration. Thus, both deconvolution and VL contribute something to the performance of the overall system.

\begin{figure}
    \centering
    \includegraphics{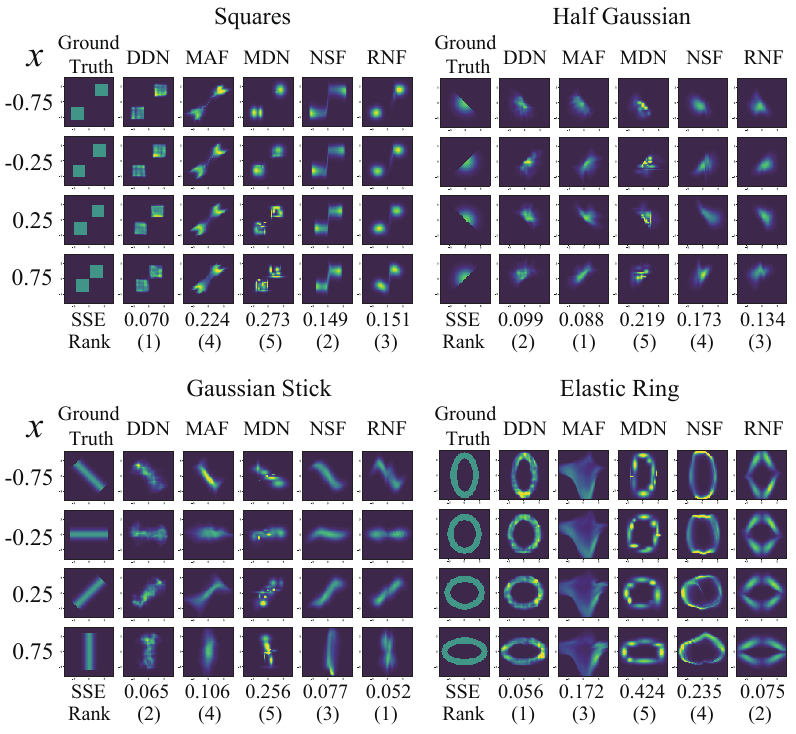} 
    \caption{Comparison of results on 2D toy tasks. Within each task, each row represents a different condition $x$, which is selected with an equal interval. The \textbf{ground truth conditional distributions} are presented in the first column, followed by the estimated distributions of DDN and other competitors, with their average SSE values over four conditions (\textit{lower is better}).
    }
    \label{fig:2D_SSE_final}
\end{figure} 

\subsection{Analysis for Hyperparameter $\beta$} 
We also studied on the impact of different $\beta$ on estimated distribution. As mentioned in \cite{Alemi2017}, $\beta$ is used to control the objective of information bottleneck, resulting in a compressed representation in latent space, which can help regularize DDN and avoid it overfitting the training data. 
Fig.~\ref{fig:beta} illustrates the larger $\beta$ can diffuse the estimated density and smoothen the function. However, an over-large $\beta$ would degrade the fidelity of an estimated density, as it puts too much weight on the homogeneity of latent variables, breaking the inter-correlation between condition and estimated density. Notice that if the available data is very sufficient, a smaller $\beta$ is suggested to avoid the over-smooth surface of a density function.

\subsection{Performance} 

\subsubsection{Toy Tasks} For each 2D toy task, we randomly generated 2000 samples from the joint probability distributions $p(x, y^{(1)}, y^{(2)})$ for training. Moreover, we trained all models over 5000 epochs. We calculated the SSE between the ground truth conditional density and the models' estimated densities. 

The results on the 2D toy task are shown in Fig.~\ref{fig:2D_SSE_final}. Moreover, the average SSE over four conditions is provided for each task. It can be observed that the DDN could adapt its estimation to diverse densities, and achieves the best average rank, which is 1.5. The closest competitor is RNF, with a score 2.25. Furthermore, there is only marginal difference the MAF and RNF are ahead of DDN at two tasks. These results verify the shapeshifting ability of DDN.

\begin{table*}
  \centering
  \small
  \begin{tabular}{cccccccc}
    \toprule
     & Fish & Concrete & Energy & Parkinsons & Temperature & Air & Skillcraft \\
    \midrule     
    MDN & -1.51±0.06 & -1.61±0.15 & -8.28±2.74 & -3.82±0.24 & -4.24±0.12 & -2.16±0.17 & -8.54±0.43\\
    
    MAF & -2.17±0.44 & -2.50±1.23 & -124.93±151.59 & -20.13±5.11 & -13.97±1.21 & -14.48±4.09 & -81.07±24.63 \\ 

    NSF & -1.38±0.14 & -1.09±0.11 & -2.87±0.33 & -1.81±0.09 & -2.95±0.13 & 0.47±0.32 & -8.68±0.26 \\    

    RNF & -1.38±0.12 & -1.71±0.25 & -19.38±12.54 & -4.01±0.76 & -7.51±1.85 & -0.81±0.77 & -26.76±6.66 \\    

    JBCENN & -6.62±0.36 & -7.19±0.36 & N/A & N/A & N/A & N/A & N/A \\    
 
    MLP & N/A & N/A & -3.48±0.12 & -4.86±0.17 & -14.01±0.13 & N/A & N/A \\    
    
    \midrule
    DDN ($\beta = 0.5$) & -1.19±0.04 & \bf{-0.58±0.14} & 0.65±0.19 & -1.37±0.04 & -1.65±0.02 & -3.36±0.25 & -4.95±0.14 \\     
    DDN ($\beta = 0.1$) & \bf{-1.11±0.08} & -1.56±0.34 & \bf{1.87±0.23} & -0.43±0.03 & -0.76±0.08 & 0.94±0.06 & -1.95±0.37 \\     
    DDN ($\beta = 0.02$) & -1.34±0.11 & -2.37±0.37 & 0.14±0.95 & \bf{-0.14±0.03} & \bf{-0.71±0.07} & 1.22±0.05 & \bf{-1.56±0.07} \\     
    DDN (no VL) & -2.24±0.18  & -3.78±0.81 & -1.56±0.81 & -0.17±0.05 & -0.84±0.07 & \bf{1.32±0.05} & -1.59±0.09 \\     
    \bottomrule
  \end{tabular}
  \caption{Comparison of log-likelihood on real-world tasks. Higher value of average log-likelihood indicates better result.}
  \label{tab:ll_std}  
\end{table*}

\subsubsection{Real World Tasks} We also compared DDN with other six CDE techniques on real world datasets. In addition, DDN with different $\beta$ values ($\beta \in \{0.5, 0.1, 0.02\}$) and DDN without VL are all reported for comparison. The models were trained independently for 10 trials on each dataset, with 3000 epochs per experiment. For each trial, we randomly split the data into a training set and test set with a ratio of 3:7, to mimic data deficiency scenario.

Table~\ref{tab:ll_std} reports the average log-likelihood and standard deviation of log-likelihood over 10 trials. The DDNs achieve the best average log-likelihood among all the methods. It should be noted that the DDN (no VL) also yields better performance, but VL further improves it. It verifies the deconvolution layers can aid DDN to adapt the real-world distributions. Notice that the DDNs prorgessively stretch their leading advantage, as the increase of difference between distributions in higher dimensional space. Moreover, DDN also achieves the lowest standard deviation on four datasets out of five. Among these results, MLP reaches the lowest standard deviation on the Energy, but DDN yields the closest results to MLP on this one. This exhibits the DDN is stabler in estimating free-form conditional distributions.

\section{Conclusion} 
\label{sec_conclusion}

In this work, we design a deconvolutional density network framework for conditional density estimation to explicitly estimate free-form PDF for continuous-domain, especially coping with the data deficiency problem. Assuming the true PDFs are commonly smooth, we use multi-layer deconvolution to hierarchically construct distributions that exhibit spatial coherence at multiple scales, to mitigate the shortage of data. In the experiment, the DDN model clearly showed the best all-around performance on all tasks.

In the future, much more work is required to unpack the features and to improve the performance of the DDN. For example, how does the deconvolution architecture (number of layers, kernel size) affect the performance? How does the dimension of the latent space depend on the dataset? Is there a better alternative to using piecewise-constant functions for creating our free-form distributions? How to accelerate the convergence of DDN, especially for high-dimensional tasks?

\section{Acknowledgments}
This work was supported by National Natural Science Foundation of China under Grant No. 61872419, No. 62072213, No. 61573166. Taishan Scholars Program of Shandong Province, China, under Grant No. tsqn201812077. We would like to take this opportunity to thank Prof. Jeff Orchard (University of Waterloo) and Prof. Bo Yang (University of Jinan) for their helpful comments on this work. We also thank the anonymous reviewers for their insightful comments and suggestions.

% \nobibliography{aaai22}


\begin{thebibliography}{36}
\providecommand{\natexlab}[1]{#1}

\bibitem[{{Alemi}, {Fischer}, and {Dillon}(2018)}]{Alemi_calibration}
{Alemi}, A.~A.; {Fischer}, I.; and {Dillon}, J.~V. 2018.
\newblock {Uncertainty in the Variational Information Bottleneck}.
\newblock \emph{arXiv e-prints}, arXiv:1807.00906.

\bibitem[{Alemi et~al.(2017)Alemi, Fischer, Dillon, and Murphy}]{Alemi2017}
Alemi, A.~A.; Fischer, I.; Dillon, J.~V.; and Murphy, K. 2017.
\newblock {Deep variational information bottleneck}.
\newblock \emph{5th International Conference on Learning Representations, ICLR
  2017 - Conference Track Proceedings}, 1--19.

\bibitem[{Ambrogioni et~al.(2017)Ambrogioni, G{\"u}çl{\"u}, Gerven, and
  Maris}]{Ambrogioni_2017}
Ambrogioni, L.; G{\"u}çl{\"u}, U.; Gerven, M.~V.; and Maris, E. 2017.
\newblock The Kernel Mixture Network: A Nonparametric Method for Conditional
  Density Estimation of Continuous Random Variables.
\newblock \emph{arXiv: Machine Learning}.

\bibitem[{Bishop(1994)}]{BISHOP_2006}
Bishop, C.~M. 1994.
\newblock Mixture density networks.
\newblock \emph{NCRG/94/004}.

\bibitem[{Dua and Graff(2017)}]{Dua:2019}
Dua, D.; and Graff, C. 2017.
\newblock {UCI} Machine Learning Repository.

\bibitem[{Durkan et~al.(2019)Durkan, Bekasov, Murray, and
  Papamakarios}]{Neural_Spline_Flows}
Durkan, C.; Bekasov, A.; Murray, I.; and Papamakarios, G. 2019.
\newblock Neural Spline Flows.
\newblock In Wallach, H.; Larochelle, H.; Beygelzimer, A.; d\textquotesingle
  Alch\'{e}-Buc, F.; Fox, E.; and Garnett, R., eds., \emph{Advances in Neural
  Information Processing Systems}, volume~32. Curran Associates, Inc.

\bibitem[{Dutordoir et~al.(2018)Dutordoir, Salimbeni, Hensman, and
  Deisenroth}]{GP_CDE}
Dutordoir, V.; Salimbeni, H.; Hensman, J.; and Deisenroth, M. 2018.
\newblock Gaussian Process Conditional Density Estimation.
\newblock In Bengio, S.; Wallach, H.; Larochelle, H.; Grauman, K.;
  Cesa-Bianchi, N.; and Garnett, R., eds., \emph{Advances in Neural Information
  Processing Systems}, volume~31. Curran Associates, Inc.

\bibitem[{Higgins et~al.(2017)Higgins, Matthey, Pal, Burgess, Glorot,
  Botvinick, Mohamed, and Lerchner}]{Higgins2017betaVAELB}
Higgins, I.; Matthey, L.; Pal, A.; Burgess, C.~P.; Glorot, X.; Botvinick, M.;
  Mohamed, S.; and Lerchner, A. 2017.
\newblock beta-{VAE}: Learning Basic Visual Concepts with a Constrained
  Variational Framework.
\newblock In \emph{ICLR}.

\bibitem[{Kingma and Welling(2019)}]{Kingma_2019_VAE}
Kingma, D.~P.; and Welling, M. 2019.
\newblock An Introduction to Variational Autoencoders.
\newblock \emph{Foundations and Trends in Machine Learning}, 12(4): 307–392.

\bibitem[{Larochelle and Murray(2011)}]{pmlr-v15-larochelle11a}
Larochelle, H.; and Murray, I. 2011.
\newblock The Neural Autoregressive Distribution Estimator.
\newblock In Gordon, G.; Dunson, D.; and Dudík, M., eds., \emph{Proceedings of
  the Fourteenth International Conference on Artificial Intelligence and
  Statistics}, volume~15 of \emph{Proceedings of Machine Learning Research},
  29--37. Fort Lauderdale, FL, USA: PMLR.

\bibitem[{Li, Reich, and Bondell(2021)}]{LI_2021}
Li, R.; Reich, B.~J.; and Bondell, H.~D. 2021.
\newblock Deep distribution regression.
\newblock \emph{Computational Statistics \& Data Analysis}, 159: 107203.

\bibitem[{Li, Liu, and Zhu(2007)}]{Li_2007}
Li, Y.; Liu, Y.; and Zhu, J. 2007.
\newblock Quantile Regression in Reproducing Kernel Hilbert Spaces.
\newblock \emph{Journal of the American Statistical Association}, 102(477):
  255--268.

\bibitem[{Papamakarios, Pavlakou, and Murray(2017)}]{MAF}
Papamakarios, G.; Pavlakou, T.; and Murray, I. 2017.
\newblock Masked Autoregressive Flow for Density Estimation.
\newblock In \emph{Proceedings of the 31st International Conference on Neural
  Information Processing Systems}, NIPS'17, 2335–2344. Red Hook, NY, USA:
  Curran Associates Inc.
\newblock ISBN 9781510860964.

\bibitem[{Rezende and Mohamed(2015)}]{Rezende_2015}
Rezende, D.; and Mohamed, S. 2015.
\newblock Variational Inference with Normalizing Flows.
\newblock In Bach, F.; and Blei, D., eds., \emph{Proceedings of the 32nd
  International Conference on Machine Learning}, volume~37 of \emph{Proceedings
  of Machine Learning Research}, 1530--1538. Lille, France: PMLR.

\bibitem[{Rothfuss et~al.(2020)Rothfuss, Ferreira, Boehm, Walther, Ulrich,
  Asfour, and Krause}]{rothfuss2020noise}
Rothfuss, J.; Ferreira, F.; Boehm, S.; Walther, S.; Ulrich, M.; Asfour, T.; and
  Krause, A. 2020.
\newblock Noise Regularization for Conditional Density Estimation.
\newblock arXiv:1907.08982.

\bibitem[{Rothfuss et~al.(2019)Rothfuss, Ferreira, Walther, and
  Ulrich}]{Jonas_2019}
Rothfuss, J.; Ferreira, F.; Walther, S.; and Ulrich, M. 2019.
\newblock {Conditional Density Estimation with Neural Networks: Best Practices
  and Benchmarks}.
\newblock Papers 1903.00954, arXiv.org.

\bibitem[{Salimans et~al.(2017)Salimans, Karpathy, Chen, and
  Kingma}]{PixelCNN++}
Salimans, T.; Karpathy, A.; Chen, X.; and Kingma, D.~P. 2017.
\newblock PixelCNN++: Improving the PixelCNN with Discretized Logistic Mixture
  Likelihood and Other Modifications.
\newblock In \emph{5th International Conference on Learning Representations,
  {ICLR} 2017, Toulon, France, April 24-26, 2017, Conference Track
  Proceedings}. OpenReview.net.

\bibitem[{Sasaki and Hyvärinen(2018)}]{sasaki_2018}
Sasaki, H.; and Hyvärinen, A. 2018.
\newblock Neural-Kernelized Conditional Density Estimation.
\newblock arXiv:1806.01754.

\bibitem[{Shao et~al.(2020)Shao, Molina, Vergari, Stelzner, Peharz, Liebig, and
  Kersting}]{Conditional_Sumproduct_Networks}
Shao, X.; Molina, A.; Vergari, A.; Stelzner, K.; Peharz, R.; Liebig, T.; and
  Kersting, K. 2020.
\newblock Conditional Sum-Product Networks: Imposing Structure on Deep
  Probabilistic Architectures.
\newblock In Jaeger, M.; and Nielsen, T.~D., eds., \emph{International
  Conference on Probabilistic Graphical Models, {PGM} 2020, 23-25 September
  2020, Aalborg, Hotel Comwell Rebild Bakker, Sk{\o}rping, Denmark}, volume 138
  of \emph{Proceedings of Machine Learning Research}, 401--412. {PMLR}.

\bibitem[{Shelhamer, Long, and Darrell(2017)}]{Shelhamer_Cov_upsam}
Shelhamer, E.; Long, J.; and Darrell, T. 2017.
\newblock Fully Convolutional Networks for Semantic Segmentation.
\newblock \emph{IEEE Transactions on Pattern Analysis \& Machine Intelligence},
  39(04): 640--651.

\bibitem[{Sohn, Lee, and Yan(2015)}]{CVAE}
Sohn, K.; Lee, H.; and Yan, X. 2015.
\newblock Learning Structured Output Representation using Deep Conditional
  Generative Models.
\newblock In Cortes, C.; Lawrence, N.; Lee, D.; Sugiyama, M.; and Garnett, R.,
  eds., \emph{Advances in Neural Information Processing Systems}, volume~28.
  Curran Associates, Inc.

\bibitem[{Strauss and Oliva(2021)}]{strauss2021arbitrary}
Strauss, R.~R.; and Oliva, J.~B. 2021.
\newblock Arbitrary Conditional Distributions with Energy.
\newblock arXiv:2102.04426.

\bibitem[{{Sugiyama} et~al.(2010){Sugiyama}, {Takeuchi}, {Suzuki}, {Kanamori},
  {Hachiya}, and {Okanohara}}]{Sugiyama_2010}
{Sugiyama}, M.; {Takeuchi}, I.; {Suzuki}, T.; {Kanamori}, T.; {Hachiya}, H.;
  and {Okanohara}, D. 2010.
\newblock {Least-Squares Conditional Density Estimation}.
\newblock \emph{IEICE Transactions on Information and Systems}, 93(3):
  583--594.

\bibitem[{Tabak and Turner(2013)}]{Tabak2013AFO}
Tabak, E.; and Turner, C. 2013.
\newblock A Family of Nonparametric Density Estimation Algorithms.
\newblock \emph{Communications on Pure and Applied Mathematics}, 66: 145--164.

\bibitem[{Tabak and Vanden-Eijnden(2010)}]{cms/1266935020}
Tabak, E.~G.; and Vanden-Eijnden, E. 2010.
\newblock {Density estimation by dual ascent of the log-likelihood}.
\newblock \emph{Communications in Mathematical Sciences}, 8(1): 217 -- 233.

\bibitem[{Takeuchi et~al.(2006)Takeuchi, Le, Sears, and Smola}]{Takeuchi_2006}
Takeuchi, I.; Le, Q.~V.; Sears, T.~D.; and Smola, A.~J. 2006.
\newblock Nonparametric Quantile Estimation.
\newblock \emph{Journal of Machine Learning Research}, 7(45): 1231--1264.

\bibitem[{Takeuchi, Nomura, and Kanamori(2009)}]{Takeuchi_2009}
Takeuchi, I.; Nomura, K.; and Kanamori, T. 2009.
\newblock {Nonparametric Conditional Density Estimation Using Piecewise-Linear
  Solution Path of Kernel Quantile Regression}.
\newblock \emph{Neural Computation}, 21(2): 533--559.

\bibitem[{{Tansey}, {Pichotta}, and {Scott}(2016)}]{Tansey_2016}
{Tansey}, W.; {Pichotta}, K.; and {Scott}, J.~G. 2016.
\newblock {Better Conditional Density Estimation for Neural Networks}.
\newblock \emph{arXiv e-prints}, arXiv:1606.02321.

\bibitem[{{Trippe} and {Turner}(2018)}]{Trippe_2018}
{Trippe}, B.~L.; and {Turner}, R.~E. 2018.
\newblock {Conditional Density Estimation with Bayesian Normalising Flows}.
\newblock \emph{arXiv e-prints}, arXiv:1802.04908.

\bibitem[{Tsybakov(2008)}]{Tsybakov_2008}
Tsybakov, A.~B. 2008.
\newblock \emph{Introduction to Nonparametric Estimation}.
\newblock Springer Publishing Company, Incorporated, 1st edition.
\newblock ISBN 0387790519.

\bibitem[{Uria, Murray, and Larochelle(2013)}]{RNADE}
Uria, B.; Murray, I.; and Larochelle, H. 2013.
\newblock RNADE: The real-valued neural autoregressive density-estimator.
\newblock In Burges, C. J.~C.; Bottou, L.; Welling, M.; Ghahramani, Z.; and
  Weinberger, K.~Q., eds., \emph{Advances in Neural Information Processing
  Systems}, volume~26. Curran Associates, Inc.

\bibitem[{Van Den~Oord et~al.(2016)Van Den~Oord, Kalchbrenner, Espeholt,
  Kavukcuoglu, Vinyals, and Graves}]{GatedPixelCNN}
Van Den~Oord, A.; Kalchbrenner, N.; Espeholt, L.; Kavukcuoglu, K.; Vinyals, O.;
  and Graves, A. 2016.
\newblock Conditional Image Generation with PixelCNN Decoders.
\newblock In Lee, D.; Sugiyama, M.; Luxburg, U.; Guyon, I.; and Garnett, R.,
  eds., \emph{Advances in Neural Information Processing Systems}, volume~29.
  Curran Associates, Inc.

\bibitem[{Van Den~Oord, Kalchbrenner, and Kavukcuoglu(2016)}]{PixelRNN}
Van Den~Oord, A.; Kalchbrenner, N.; and Kavukcuoglu, K. 2016.
\newblock Pixel Recurrent Neural Networks.
\newblock In \emph{Proceedings of the 33rd International Conference on
  International Conference on Machine Learning - Volume 48}, ICML'16,
  1747–1756. JMLR.org.

\bibitem[{Winkler et~al.(2019)Winkler, Worrall, Hoogeboom, and Welling}]{CNFs}
Winkler, C.; Worrall, D.; Hoogeboom, E.; and Welling, M. 2019.
\newblock Learning Likelihoods with Conditional Normalizing Flows.
\newblock arXiv:1912.00042.

\bibitem[{Xiao, Yan, and Amit(2019)}]{xiao2019method}
Xiao, Z.; Yan, Q.; and Amit, Y. 2019.
\newblock A Method to Model Conditional Distributions with Normalizing Flows.
\newblock arXiv:1911.02052.

\bibitem[{Yu et~al.(2017)Yu, Gong, Zhong, and Shan}]{DCGAN_YU}
Yu, Y.; Gong, Z.; Zhong, P.; and Shan, J. 2017.
\newblock Unsupervised Representation Learning with Deep Convolutional Neural
  Network for Remote Sensing Images.
\newblock In Zhao, Y.; Kong, X.; and Taubman, D., eds., \emph{Image and
  Graphics}, 97--108. Cham: Springer International Publishing.
\newblock ISBN 978-3-319-71589-6.

\end{thebibliography}
\end{document}